# CUICurate: A GraphRAG-based Framework for Automated Clinical Concept Curation for NLP applications

Victoria BLAKE, MSc[a,b], Jamie NOVAK, MD[e], Mathew MILLER, MBChB[c,d,f], Sze-Yuan OOI, MD[b,e,f] and Blanca GALLEGO, PhD[a]

[a] *Centre for Big Data Research in Health, University of New South Wales, Sydney, New South Wales, Australia*

[b] *Eastern Heart Clinic, Prince of Wales Hospital, Randwick, NSW, Australia*

[c] *NSW Ambulance Aeromedical Operations, Bankstown Helicopter Base, Sydney*

[d] *Department of Anaesthesia, Saint George Hospital, Kogarah, New South Wales, Australia*

[e] *Department of Cardiology, Prince of Wales Hospital, South Eastern Sydney Local Health District, Randwick, NSW, Australia*

[f] *School of Clinical Medicine, University of New South Wales, Sydney, New South Wales, Australia*

**Corresponding Author:** Victoria Blake, Centre for Big Data Research in Health, UNSW, Sydney NSW, Australia, v.blake@unsw.edu.au,

## ABSTRACT:

**Background**: Clinical named entity recognition tools commonly map free text to Unified Medical Language System (UMLS) Concept Unique Identifiers (CUIs). For many downstream tasks, however, the clinically meaningful unit is not a single CUI but a *concept set* comprising related synonyms, subtypes, and associated concepts. Constructing these sets is labour-intensive, inconsistently performed, and poorly supported by existing tools.

**Methods** We present CUICurate, a graph-based retrieval-augmented generation (GraphRAG) framework for automated UMLS concept set curation. A UMLS knowledge graph (KG) was constructed and embedded for semantic retrieval. Candidate CUIs were retrieved using graph-based expansion and then filtered and classified using large language models (GPT-5 and Qwen3-32B). The framework was evaluated on five lexically heterogeneous clinical concepts against a manually curated concept sets and gold-standard concept sets. **Results** CUICurate produced substantially larger and more complete concept sets than the manual benchmarks. A single retrieval configuration across concepts achieved high recall of definitive concepts with manageable candidate sets. GPT-5 outperformed manual curation for all concepts and retained at least 95% of definitive gold-standard CUIs, while Qwen3-32B achieved comparable but slightly lower performance. Many missed concepts were not observed in 10,000 MIMIC-III notes. CUICurate infrastructure and end-to-end processing was inexpensive and stable across runs. **Conclusions** CUICurate offers a scalable, reproducible and cost-efficient approach for generating clinician-reviewable UMLS concept sets tailored to clinical natural language processing and phenotyping applications.

## INTRODUCTION

Clinical natural language processing (NLP) tools often map clinical terms to Concept Unique Identifiers (CUIs) from the Unified Medical Language System (UMLS).[1] The UMLS contains more than 4 million biomedical concepts drawn from over 180 source vocabularies, giving broad coverage of the many ways clinicians describe conditions, findings and interventions, making it a powerful foundation for clinical concept normalisation.[1, 2] However, its breadth and inclusivity also create challenges as concepts can overlap, use ambiguous wording, or be linked inconsistently through parent-child or synonymous relationships.[1] As a result, it can be difficult to reliably identify all textual expressions that indicate a given clinical concept when extracting information from text.

For many downstream tasks, the clinically meaningful unit is not a single concept, but a set of related concepts encompassing synonyms, subtypes and supertypes. For example, identifying conditions such as heart failure requires grouping individual CUIs such as *chronic heart failure*, *congestive heart failure* or *left-sided heart failure* into a list of related concepts that together represent a single clinical idea. These lists, often referred to as concept sets or value sets, define which CUIs should be treated as evidence for the target concept, and is a key process if concepts from text are to be comparable and reproducible across datasets.[3] This is a particular challenge in NLP applications, where concept normalisation tools may map the same clinical expression to different CUIs depending on context, wording or tool behaviour as illustrated in Figure 1.

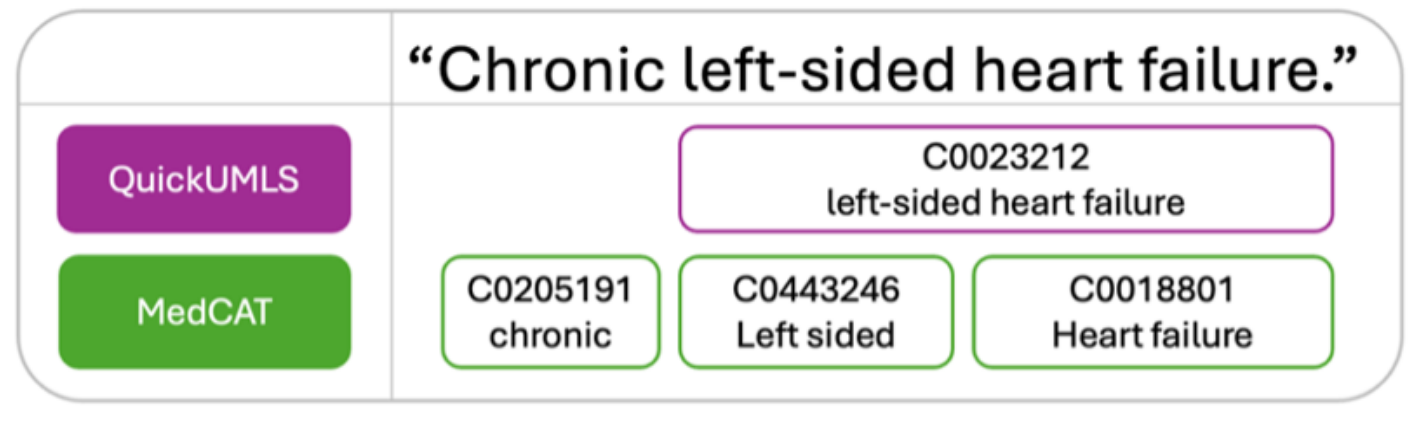

*Figure 1. **Example variation in concept normalisation across NLP tools**. The same clinical phrase is mapped to different CUIs by the QuickUMLS[4] and MedCAT[5] tools due to differences in tool behaviour*

Constructing such concept sets is a manual and labour-intensive process.[6] It requires specialist clinical judgement, iterative searching and careful review to decide which concepts are relevant and which should be included. Existing tools offer only partial support. The UMLS Metathesaurus Browser[7] enables manual exploration of CUIs, but offers limited support for collecting groups of related concepts as a coherent concept set. The Observational Health Data Sciences and Informatics (OHDSI) Athena[8] platform is purpose built for constructing concept sets and provides more advanced search and hierarchical expansion across multiple vocabularies, but still relies heavily on manual curation. Moreover, Athena concept sets are designed primarily for structured, coded data and do not operate directly on UMLS CUIs, requiring additional mapping steps when used in NLP pipelines that extract concepts from free text. As a result, concept sets remain time-consuming to build and difficult to maintain.

Some vocabularies within the UMLS, such as Systematized Nomenclature of Medicine - Clinical Terms (SNOMED CT), provide hierarchical relationships that can be used to identify related concepts. While these hierarchies are useful, relying on hierarchical expansion alone is insufficient for constructing clinically meaningful concept sets. Relevant variants may appear at different levels of the hierarchy or outside a single branch, while broader parent concepts can introduce clinically inappropriate terms. Also, clinically relevant variants are not always directly connected through parent-child relationships, reflecting differences in how concepts are organised across source vocabularies. Figure 2 illustrated how CUIs relevant to a single clinical concept may be distributed across the UMLS graph, highlighting the limitations of hierarchy-based retrieval.

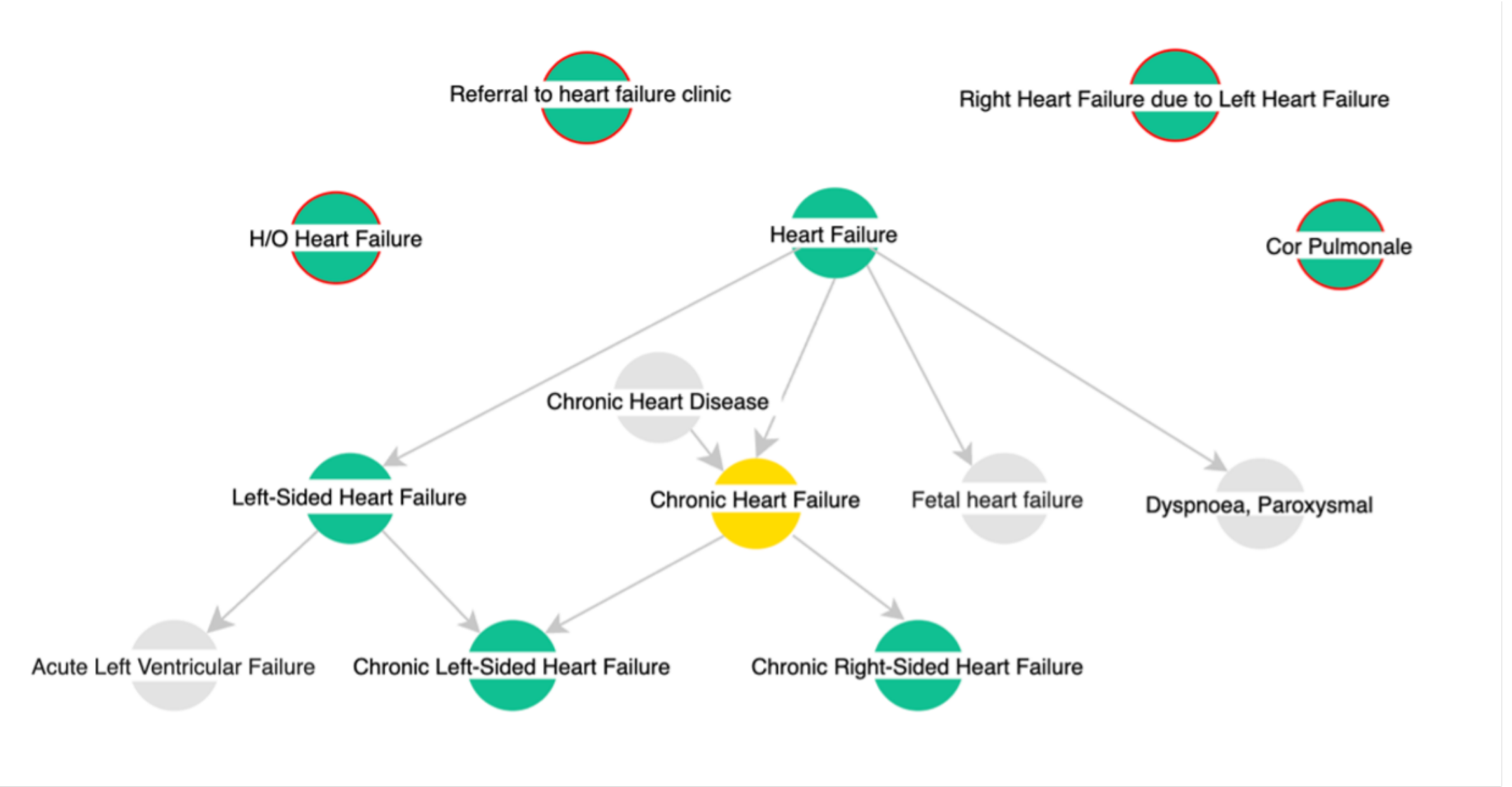

*Figure 2. **Example UMLS subgraph centred on chronic heart failure**. The target concept is shown in yellow, clinically relevant variants in green, and related but clinically inappropriate concepts in grey. Nodes outlined in red represent relevant CUIs that are not directly connected to the target concept and would be missed by hierarchy-based retrieval. Arrows represent hierarchical relationships (parent to child) encoded in the UMLS.*

The automation of concept set construction has received little direct attention. Large language models (LLMs) have demonstrated strong performance on a range of biomedical reasoning tasks, including assessing semantic relevance and handling clinical nuance,[9] making them well suited to judging whether concepts are meaningfully related even when they are not hierarchically connected. However, applying LLMs directly to the full UMLS is impractical due to its size, cost, and the risk of overwhelming the model with irrelevant information. Moreover, approaches that rely on flat lexical or embedding-based retrieval alone ignore the rich relational structure of biomedical terminologies. As a result, LLMs require a constrained and clinically relevant search space in order to be applied effectively and efficiently to concept curation tasks.

Retrieval-augmented generation (RAG)[10] provides a mechanism for constraining LLM inputs by retrieving a subset of relevant items to include as context. Graph-based variants of RAG known as GraphRAG[11] extend this idea by performing retrieval over a graph structure rather than a flat document collection, enabling both semantic similarity and explicit relationships to be used during retrieval. This approach is particularly well suited to knowledge sources such as the UMLS, which is inherently graph-structured and encodes relationships between concepts across multiple vocabularies. In this study, we propose CUICurate, a GraphRAG-based framework that combines graph-based retrieval over the UMLS with LLM-based filtering and classification to automate the construction of clinically meaningful concept sets. By first retrieving a focused subgraph of candidate CUIs and then applying LLM reasoning to assess relevance and contextual certainty, this approach aims to reduce manual effort while producing curated clinically coherent concept sets that can be rapidly reviewed by clinicians and applied in NLP pipelines that map text to UMLS CUIs.

**Related Work**

Several initiatives have standardised clinical concepts through curated codes lists, including OHDSI concept sets and the National Library of Medicine's Value Set Authority Centre (VSAC).[3] These resources support reproducible analysis of structured data but are typically tailored to specific studies and built around terminologies such as SNOMED CT or International Classification of Diseases (ICD), limiting direct applicability to NLP pipelines based on UMLS CUIs.[3, 12]

Prior work on phenotype concept set construction includes supervised approaches that learns from existing curated phenotypes.[6] However, these methods rely on prior phenotype definitions and do not target UMLS-based NLP workflows. Large language models have also been applied to concept normalisation, where detected text mentions are mapped to the most appropriate UMLS concept. Dobbins et al. improved mention-level normalisation by combining existing tools with LLM-based synonym generation and candidate pruning.[13]

Graph-based retrieval has similarly been explored in biomedical question answering, where GraphRAG provides structured context for LLM reasoning.[14]

However, the construction of reusable UMLS concept sets for downstream NLP applications remains largely manual and disconnected from concept extraction pipelines.

# METHODS

## Data Sources

The UMLS Rich Release Format (RRF) files were used to construct the knowledge graph, including concept names (MRCONSO), relationships (MRREL), definitions (MRDEF), and semantic types (MRSTY). Three source vocabularies with well-defined relational structure were included: SNOMED CT–US, the NCI Thesaurus (NCIt), and Medical Subject Headings (MeSH).

## Target Concepts

Five lexically heterogenous concepts representing a diverse range of clinical conditions and also known to have extensive sub-types or varied ways of being expressed were selected to develop and test CUICurate, which included *chronic heart failure (CHF), ischaemic stroke (IS)*, *fluid overload (FO)*, *left ventricular systolic dysfunction (LVSD)* and *poor mobility (PM*).

## Knowledge Graph Construction, Node Embedding & Indexing

A bi-directed knowledge graph was constructed from UMLS relationships, with CUIs represented as nodes and semantic relationships encoded as edges. Node attributes included synonyms, definitions and semantic types. Each node was embedded using three models representing general-purpose, domain-specific biomedical, and open-source general purpose embeddings: ***text-embedding-3-large*** (OpenAI) a high-performing general-purpose closed model for semantic similarity and retrieval tasks[15], **SapBERT** a BERT-based model trained using UMLS synonym pairs[16], and **e5-large-v2** an open-source general-purpose embedding model with strong performance across diverse embedding clinical and biomedical benchmarks.[17, 18] Embeddings were indexed using the FAISS cosine similarity algorithm (*IndexFlatIP)*.[19]An overview of the CUICurate framework is shown in Figure 3.

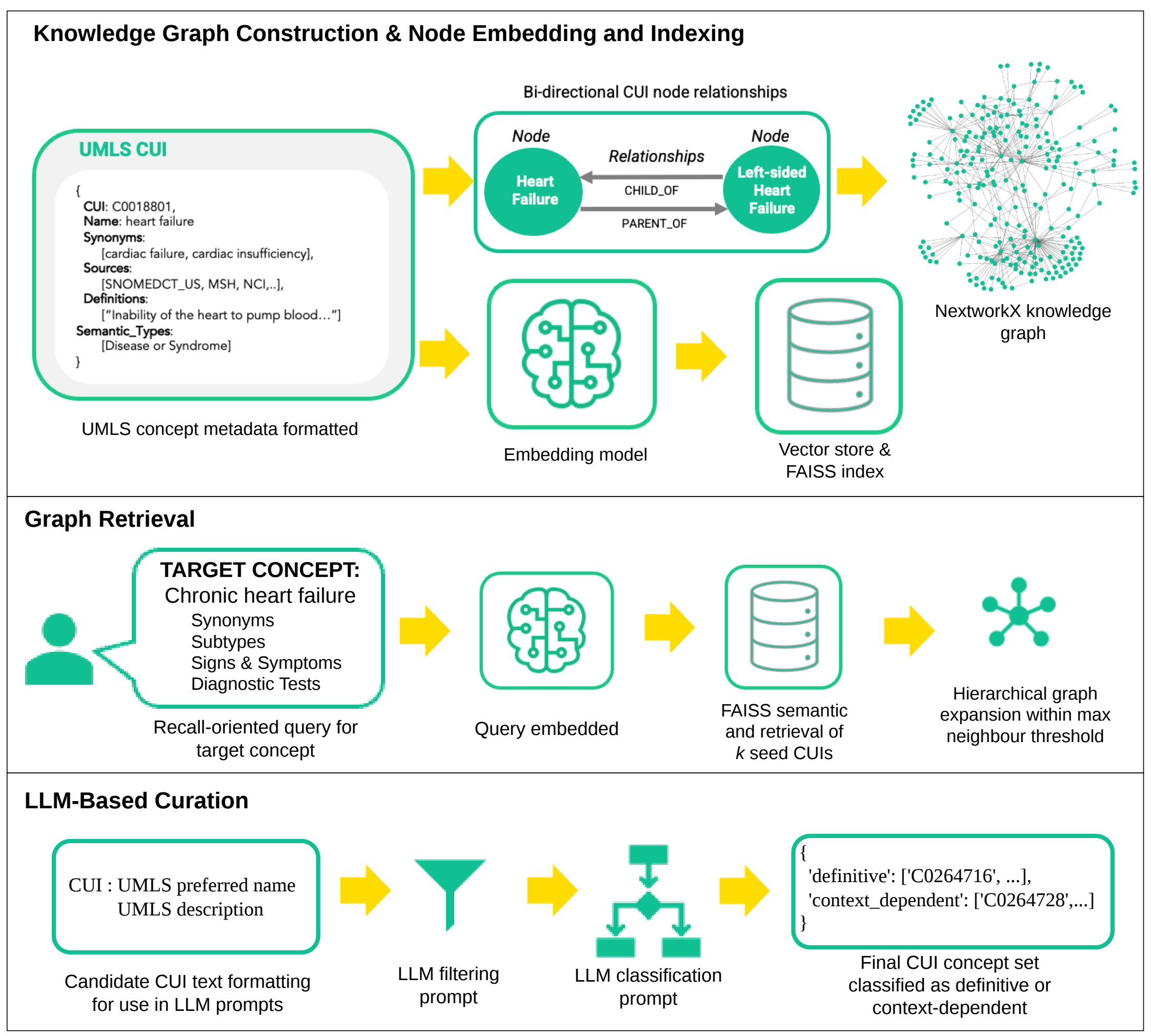

*Figure 3.* ***Overview of CUICurate, a GraphRAG-based framework for automated UMLS concept set curation****.* *UMLS concepts were represented as a knowledge graph. Each concept forms one node within the graph with UMLS relationships forming the node edges. Each node was embedded using an embedding and indexed using FAISS model for semantic retrieval. For a given target concept, a recall-oriented descriptionis embedded using the same*

*embedding model which is used to query the FAISS index, retrieving k nearest neighbour seed CUIs from the graph, which are expanded through hierarchical graph relationships to a threshold of semantic max neighbour CUIs, to form a focused candidate set. This CUI candidate set is then filtered for relevance and classified by an LLM chat-model into definitive and context-dependent indicators of the target concept. Target concept descriptions can be found in the supplementary materials. Abbreviations: UMLS, Unified Medical Language System; CUI, Concept Unique Identifier; FAISS, Facebook AI Similarity Search; LLM, Large Language Model*

### Graph Retrieval

Graph retrieval was designed to identify a high-recall but computationally manageable set of candidate CUIs for subsequent LLM curation. For each target concept, a recall-oriented description was embedded and used to retrieve the top-*k* semantically similar CUIs. Recursive child expansion along with one-hop parent and synonymous graph expansion (***hops***) incorporated hierarchical related concepts. Candidate sets were constrained using semtantic type filters and neighbourhood limits (**max_neighbors**) to maintain managable prompt sizes. Additional details and parameter definitions are provided in the Supplementary materials.

### LLM-Based Curation

Retrieved CUIs were curated using a two-stage LLM workflow. First, CUIs were filtered to retain concepts that indicate, or strongly suggest, the target phenotype. Second, included CUIs were classified as ***definitive*** (explicitly or strongly indicating the target concept), or ***context-dependent**,* meaning the concept may indicate the target concept, but requires additional clinical information for confirmation.

Two chat models were compared: GPT-5 accessed through the OpenAI API, and Qwen3-32B[20], an open-source LLM that was deployed locally. A 4-bit quantized version of Qwen3-32B was used to enable inference on a single high-memory GPU. Both models were run five times for each target concept to assess variability in LLM outputs.

Prompt templates and implementation details are provided in the Supplementary materials.

**Benchmark and Gold-Standard Concept Sets**

In the absence of existing benchmark concept sets for the selected target concepts, manual concept sets (***M***) were collected by two clinicians using the UMLS Metathesaurus Browser and adjudicated by a third clinician. Discrepancies were resolved by a third clinician, and agreement between the independently collected sets was assessed using Jaccard and overlap coefficients.

Because exhaustive review of all potentially relevant CUIs in the UMLS Metathesaurus was infeasible, an adjudicated gold-standard concept set ($\boldsymbol{G_a}$) was constructed using pooled relevance assessment[21], a standard methodology developed in information retrieval evaluation to create reference standards from the union of candidates retrieved by multiple systems. This approach has been widely used in Text REtrieval Conference (TREC) evaluations and has been shown to remain robust for evaluating modern neural retrieval systems[22].

Candidate CUIs retrieved using the three embedding models described above, together with graph traversal using hierarchical and hop neighbourhood expansion, were combined to form a comprehensive candidate retrieval pool which were independently annotated by two clinicians. Disagreements were resolved by consensus. Interrater agreement was assessed using percent agreement and Cohen's κ. The resulting adjudicated set of included CUIs formed the gold-standard concept set ($\boldsymbol{G_a}$) used for downstream evaluation.

**Evaluation Framework**

Each stage of the pipeline was evaluated separately using concept sets accessible to that stage (i.e., the method's retrieval or inclusion space) to ensure fair comparisons. Retrieval was evaluated against the adjudicated gold-standard concept sets ($\boldsymbol{G_a}$), using recall as the primary optimisation target.

Filtering Performance for both the LLMs and the manual benchmark (***M***) was evaluated against the subset of $G_a$ available in each run's retrieved candidate set using precision, recall and F1-score. Preservation of definitive concepts was evaluated using the same metrics, with the reference standard restricted to the subset of $G_a$ CUIs classified as ***definitive***.

Additional evaluation methodological details are provided in the supplementary materials.

### Clinical Text Observability Analysis

To assess the practical significance of missed concepts, CUIs missed during retrieval and filtering were evaluated for occurrence in a sample of 10,000 MIMIC-III clinical notes sampled across multiple document types using QuickUMLS. A similarity threshold of 0.7, score-based overlap resolution, and a 10-token matching window was used. For each target concept, gold-standard CUIs missed at graph retrieval and after LLM filtering were classified as observed or not observed in the sampled notes. This analysis was intended as a pragmatic assessment of whether recall losses affected CUIs appearing in the sampled clinical text.

CUICurate is available as a freely available python-based package which can be accessed through the CUICurate GitHub repository (https://github.com/vickyblake/CUICurate). Source code, prompt templates, and adjudicated gold-standard concept sets are publicly available in the repository.

## RESULTS

## Knowledge Graph

The UMLS graph constructed from SNOMED CT, MeSH and NCI vocabularies comprised 924,211 nodes and 1,429,114 edges. After restricting the graph to the semantic types used in the GraphRAG pipeline, the working subgraph comprised 144,646 nodes and 223,043 edges. In this restricted graph, the median node degree increased to 2 (IQR 1-3), and 8.2% of nodes were isolated.

## Manual and gold-standard concept sets

Manual concept sets (*M*) identified through the UMLS browser were substantially smaller than the pooled gold-standard sets ($G_a$), ranging from 30-130 CUIs compared with 163-675 CUIs (Table 1). The largest $G_a$ sets were observed for IS and PM, reflecting the broad and heterogenous ways in which these concepts can be represented in clinical terminology. CHF had the highest proportion of definitive concepts (69.6%), and fluid overload had the highest proportion of context-dependent concepts (92.8%). Agreement between clinicians on inclusion decisions within the pooled $G_a$ candidate sets was moderate to high, with percent agreement between 73.7% to 94.5% and Cohen's κ from 0.35–0.87).

*Table 1.* ***Manual and adjudicated gold-standard concept set composition and annotator agreement statistics****.* *Manual concept sets* ***(M)*** collected directly from the UMLS Metathesaurus browser. *Adjudicated gold-standard* ***($G_a$)*** *sets constructed by adjudicating the pooled union of candidate CUIs retrieved using three embedding models and graph expansion strategies. Agreement statistics are only shown for the* ***($G_a$)*** *concept sets, for which both clinicians independently reviewed the same pooled candidate CUIs. Because the* ***(M)*** *concept sets were generated through independent open-ended UMLS browsing rather than assessment of a fixed candidate set, agreement was summarized using Jaccard and overlap coefficients which are provided in the supplementary materials.*

*Abbreviations: CHF, chronic heart failure; FO, fluid overload; IS, ischaemic stroke; LVSD, left ventricular systolic dysfunction; PM poor mobility; Context-dep, context-dependent.*

| | | Concept set classifications | | | Agreement Statistics | | | |
|---|---|---|---|---|---|---|---|---|
| | | | | | Inclusion | | Category | |
| **Source** | **Concept** | **Total n** | **Definitive n (%)** | **Context-dep n (%)** | **IAA (%)** | **κ** | **IAA (%)** | **κ** |
| **Manual concept sets (*M*)** | **CHF** | 98 | 48 (49.0%) | 50 (51.0%) | | | | |
| | **FO** | 30 | 15 (50.0%) | 15 (50.0%) | | | | |
| | **IS** | 130 | 79 (60.8%) | 51 (39.2%) | | | | |
| | **LVSD** | 55 | 23 (41.8%) | 32 (58.2%) | | | | |
| | **PM** | 92 | 28 (30.4%) | 64 (69.6%) | | | | |
| **Gold-standard adjudicated concept sets (*Ga*)** | **CHF** | 163 | 114 (69.9%) | 49 (30.1%) | 82.3% | 0.35 | 67.5% | 0.31 |
| | **FO** | 321 | 23 (7.2%) | 298 (92.8%) | 90.2% | 0.54 | 82.4% | 0.52 |
| | **IS** | 675 | 403 (59.7%) | 273 (40.4%) | 93.9% | 0.87 | 85.7% | 0.66 |
| | **LVSD** | 287 | 51 (17.8%) | 236 (82.2%) | 94.5% | 0.76 | 75.3% | 0.47 |
| | **PM** | 655 | 355 (54.2%) | 300 (45.8%) | 73.7% | 0.39 | 73.6% | 0.43 |

## Graph Retrieval Performance

Retrieval performance varied substantially across target concepts (Figure 4). CHF showed the strongest retrieval performance, whereas FO and PM were more challenging to retrieve comprehensively. Across most settings, text-embedding-3-large outperformed SapBERT and

e5-large-v2 although differences between models were smaller for the two disease concepts CHF and IS. Recall gains diminished for most concepts beyond approximately 500 seed CUIs. Definitive CUIs was generally retrieved with higher recall at smaller values of *k* and fewer retrieved CUIs, particularly for FO and LVSD, where the definitive subsets were relatively small (23 and 51 CUIs, respectively).

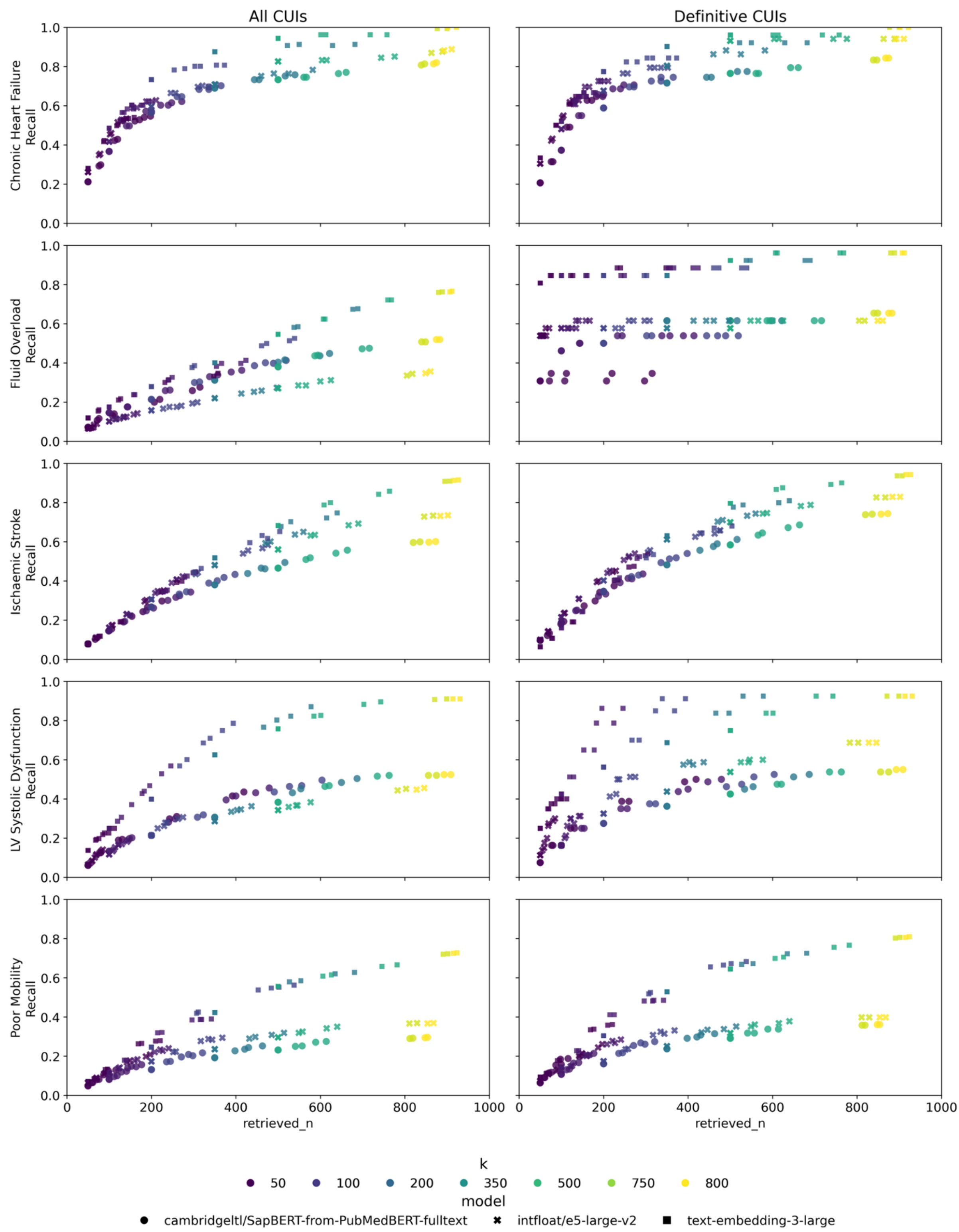

*Figure 4.* ***Retrieval performance across embedded models and graph retrieval settings.*** *Scatter plots show recall of adjudicated gold-standard concept sets (left column, all CUIs; right column, definitive CUIs only) verses the number of CUIs retrieved for five target concepts. Each point represents a unique retrieval configuration varying the number of semantically retrieved seed CUIs (k=50-800; colour scale), embedding model (marker shape), plus 0 or 1*

*expansion hops and max neighbour thresholds (50-1000). Across most concepts, text-embedding-3-large achieved the highest recall, and recall gains diminished beyond approximately 500 seed CUIs. Definitive CUIs were generally retrieved with higher recall than the full gold sets. Abbreviations: CUI, concept unique identifier; CHF, chronic heart failure; FO, fluid overload; IS, ischaemic stroke; LVSD, left ventricular systolic dysfunction; PM, poor mobility.*

To assess to feasibility of achieving complete recall, additional analyses were conducted using semantic-only retrieval and unrestrained (no max neighbours) semantic retrieval with recursive child expansion. However, these approaches often became computationally impractical for downstream LLM filtering, particularly for broad and semantically diffuse target concepts. Semantic-only retrieval alone required tens of thousands of CUIs to achieve complete recall, particularly for FO and PM (Supplementary Table n). Unrestricted child expansion (Table 2) substantially improved recall but frequently produced large increases in candidate set size and requiring multiple levels of hierarchical traversal. Using 500 seed CUIs, FO recall increased from 0.55 to 0.83, but expanded the candidate set to 2,547 CUIs, while PM recall increased from 0.44 to 0.66 but with 2,056 retrieved CUIs.

***Table 2. Semantic retrieval and unrestricted child expansion recall and number of retrieved CUIs across target concepts**. For each target concept, the maximum achievable recall was calculated using the top-k semantically retrieved seed concept unique identifiers (CUIs) and unrestrained child expansion through child hierarchical relationships. Seed recall represents the recall achievable with the seeds alone, and seed + child recall represents the recall achieved through child expansion from those seeds. Min child depth represents the number of child traversals needed to reach the maximum seed + child recall. CUIs retrieved represents the number of candidate CUIs returned from this unbounded seed + child expansion. Abbreviations: CHF, chronic heart failure; FO, fluid overload; IS, ischaemic stroke; LVSD, left ventricular systolic dysfunction; PM poor mobility.*

| | | Seeds (k) | | | | | |
|---|---|---|---|---|---|---|---|
| **concept** | **metric** | **250** | **500** | **750** | **800** | **1000** | **1500** |
| **CHF** | seed recall | 0.77 | 0.94 | 0.99 | 1.00 | 1.00 | 1.00 |
| | seed + child recall | 0.83 | 0.96 | 0.99 | 1.00 | 1.00 | 1.00 |

| | min child depth | 2 | 1 | 0 | 0 | 0 | 0 |
|---|---|---|---|---|---|---|---|
| | CUIs retrieved | 529 | 1040 | 750 | 800 | 1000 | 1500 |
| **FO** | seed recall | 0.36 | 0.55 | 0.68 | 0.70 | 0.78 | 0.83 |
| | seed + child recall | 0.76 | 0.83 | 0.89 | 0.89 | 0.90 | 0.93 |
| | min child depth | 7 | 3 | 4 | 4 | 4 | 4 |
| | CUIs retrieved | 2472 | 2547 | 3981 | 4038 | 6129 | 7909 |
| **IS** | seed recall | 0.33 | 0.60 | 0.79 | 0.81 | 0.89 | 0.95 |
| | seed + child recall | 0.67 | 0.85 | 0.95 | 0.95 | 0.98 | 0.99 |
| | min child depth | 5 | 4 | 4 | 4 | 4 | 3 |
| | CUIs retrieved | 1037 | 1469 | 1841 | 1915 | 2082 | 2812 |
| **LVSD** | seed recall | 0.44 | 0.66 | 0.80 | 0.82 | 0.87 | 0.92 |
| | seed + child recall | 0.70 | 0.87 | 0.92 | 0.93 | 0.94 | 0.96 |
| | min child depth | 3 | 3 | 2 | 2 | 2 | 1 |
| | CUIs retrieved | 497 | 1314 | 1831 | 2706 | 3499 | 2926 |
| **PM** | seed recall | 0.25 | 0.44 | 0.58 | 0.60 | 0.70 | 0.74 |
| | seed + child recall | 0.53 | 0.66 | 0.76 | 0.77 | 0.79 | 0.82 |
| | min child depth | 4 | 6 | 3 | 3 | 3 | 3 |
| | CUIs retrieved | 1129 | 2056 | 2344 | 2389 | 3319 | 4686 |

Based on these analyses, A single operating point was selected using text-embedding-3-large, 500 seed CUIs, one-hop expansion, and a neighbourhood limit of 1,000 generated approximately 750 candidate CUIs per concept while maintaining high recall (Table 3). Overall recall ranged from 0.66 to 0.96, and definitive recall exceeded 0.90 for all concepts except PM.

***Table 3. Selected GraphRAG retrieval settings and retrieval performance at those settings across target concepts.*** *The same retrieval settings were used across concepts which were text-embedding-3-large model embeddings, k=500 seed CUIs, max neighbors=1000, and one-hop graph expansion. Overall recall was calculated*

*against the full adjudicated gold-standard sets ($G_a$), while definitive recall was calculated using only $G_a$ CUIs classified as definitive indicators of the target concept. Abbreviations: CUI, concept unique identifier; CHF, chronic heart failure; FO, fluid overload; IS, ischaemic stroke; LVSD, left ventricular systolic dysfunction; PM poor mobility.*

| Concept | CUIs retrieved | Overall Recall | Definitive Recall |
|---|---|---|---|
| CHF | 758 | 0.96 | 0.96 |
| FO | 768 | 0.72 | 0.96 |
| IS | 764 | 0.85 | 0.90 |
| LVSD | 743 | 0.89 | 0.92 |
| PM | 782 | 0.67 | 0.77 |

## Filtering Performance

Figure 5 summarises filtering performance. GPT-5 achieved the highest recall and F1-scores for both the full gold-standard concept sets ($G_a$) and the subset of definitive CUIs. Qwen3-32B consistently outperformed *M* set construction in recall but generally underperformed GPT-5, particularly for FO and LVSD.

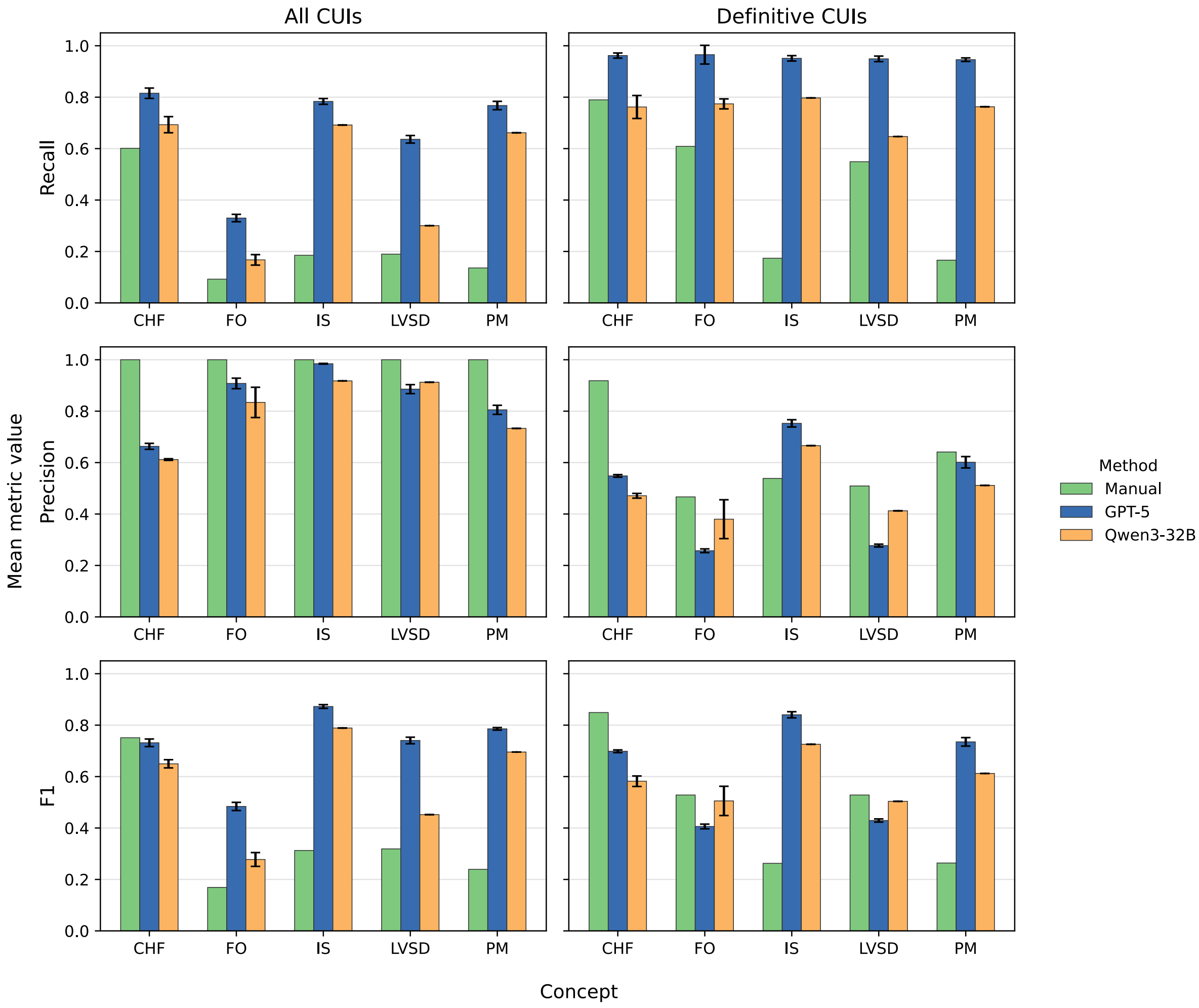

*Figure 5.* ***LLM filtering performance.*** *Mean recall, precision and F1-score across five runs per concept for GPT-5 and Qwen3-32B, compared with manual concept set collection (manual). Performance is shown for the full adjudicated gold-standard concept sets (left column, All CUIs) and for the subset of definitive CUIs only (right column, Definitive CUIs). Error bars represent the standard deviation in performance. LLM outputs evaluated against the adjudicated gold-standard concept sets using CUIs available in each methods retrieved candidate set. GPT-5 achieved the highest recall and F1-scores across most concepts. Recall was substantially higher when evaluation was restricted to definitive CUIs with GPT-5 retaining at least 95% of definitive phenotype-defining concepts across the five target concepts. Abbreviations: CUI, concept unique identifier; CHF, chronic heart failure; FO, fluid overload; IS, ischaemic stroke; LVSD, left ventricular systolic dysfunction; PM, poor mobility. Full performance metrics can be found in the Supplementary materials.*

For the full $G_a$ concept sets, GPT-5 recall ranged from 0.33 for FO to 0.82 for CHF, with particularly strong performance for IS (0.78), PM (0.77) and LVSD (0.64) in comparison to $M$ concept sets which in achieved markedly lower recall for all concepts except CHF, ranging from 0.09 for FO to 0.60 for CHF. Qwen3-32B showed intermediate performance, with recall ranging from 0.17 to 0.69.

Recall was substantially higher when evaluation was restricted to definitive CUIs. GPT-5 achieved recall of 0.95 or greater for all five concepts, demonstrating near-complete retention of the most clinically specific concepts despite more modest recall for the full $G_a$ concept sets. Qwen3-32B also maintained substantially better recall for definitive CUIs than for the complete concept sets, ranging 0.65-0.80. *M* sets showed substantially lower inclusion of definitive concepts for IS (0.17) and PM (0.17), both of which had the highest number of definitive CUIs in the $G_a$ concept sets (403 and 355, respectively).

Precision remained high for both LLMs when evaluated against the full $G_a$ concept sets, exceeding 0.80 for all concepts except CHF for GPT-5 and CHF and PM for Qwen3-32B. Precision was lower when evaluation was restricted to definitive $G_a$ concept set CUIs as all retained CUIs, including context-dependent concepts, were treated as positive predictions.

Overall, these findings demonstrate that LLM-based filtering substantially improved concept set completeness compared with conventional manual curation while maintaining acceptable precision.

**Error Analysis and Clinical Observability of Missed Concepts**

Figure 6 shows that many of the $G_a$ CUIs missed by CUICurate (text-embedding-3-large query retrieval plus GPT-5 filtering and classification) were not observed in the sampled MIMIC-III notes, particularly among context-dependent concepts. In contrast, definitive concepts observed in the clinical text were much less likely to be missed, particularly for FO and LVSD concepts.

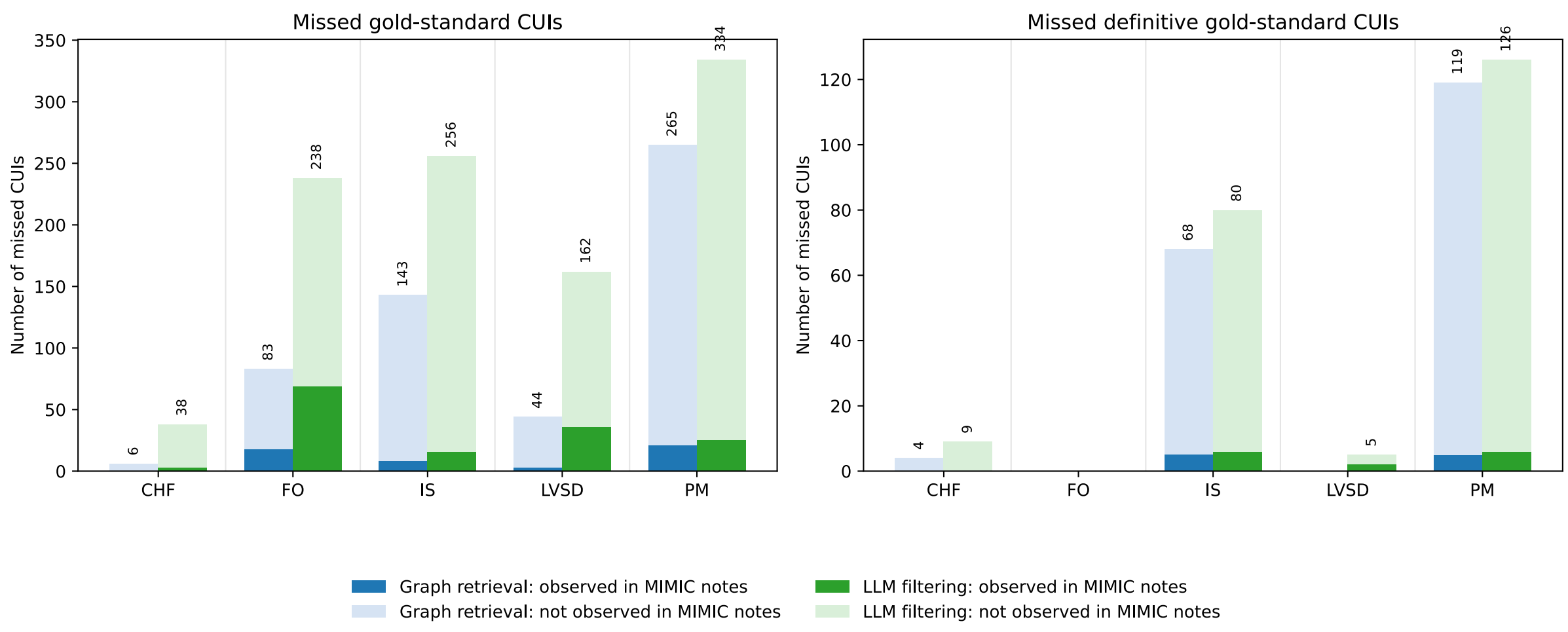

*Figure 6.* ***Clinical observability of gold-standard CUIs missed during graph retrieval and filtering stages****. Plot shows the number of adjudicated gold-standard CUIs missed at the graph retrieval stage and after the final LLM filtering stage using GPT-5 across five target concepts. Missed CUIs are stratified according to whether they were observed in a sample of 10,000 MIMIC-III clinical notes processed using QuickUMLS. The left panel shows all missed gold-standard CUIs, while the right panel shows only CUIs classified as definitive indicators of the target concept. Abbreviations: CUI, concept unique identifier; CHF, chronic heart failure; FO, fluid overload; IS, ischaemic stroke; LVSD, left ventricular systolic dysfunction; PM, poor mobility.*

Qualitative review of the concept sets demonstrated CUICurate expanded the manually curated concept sets with clinically meaningful terms found in the MIMIC-III notes and absent from the manually collected UMLS browser concept sets (*M*). Examples include New York Heart Association functional classes and cor pulmonale for CHF; generalised oedema for FO; chronic systolic heart failure for LVSD; watershed and subcortical infarct for ischaemic stroke; and functional descriptors such as unable to stand up and unable to transfer from chair to bed for PM. CUICurate also captured numerous systematic terminology variants, including anatomically specific stroke subtypes (e.g. infarction due to occlusion of individual cerebral arteries) and detailed mobility descriptors (e.g., difficulty walking up stairs, walking on slopes, and transferring).

Residual omissions were concentrated among a small number of interpretable categories. Some reflected prompt-specific decisions, such as exclusion of acute heart failure terms after instructions intended to exclude transient acute cardiac failure. Others involved generic but clinically important descriptors lacking explicit phenotype specificity, anatomically specific vascular terms not explicitly represented in the retrieval query (e.g., vertebral and basilar artery occlusion), and large families of closely related ontology variants, particularly among physiotherapy and occupational therapy terms for poor mobility. In addition, several missed poor mobility concepts represented neurological conditions commonly associated with impaired mobility, such as motor neuron disease, cerebral palsy, and multiple sclerosis, where mobility limitation is frequent but not explicitly asserted. Together, these findings suggest that remaining omissions were largely attributable to predictable semantic and ontological complexities rather than systematic failure to identify core phenotype-defining concepts. Full lists of the gold concept sets and their presence in the manual, retrieval, filtered and mimic sets are provided in the Supplementary materials.

### Compute vs. Human Resources

One-time UMLS embedding with text-embedding-3-large cost US$8.20 and required just over two hours. Graph retrieval was fast and incurred negligible cost. GPT-5 filtering and classification cost approximately $0.88 USD per query and required about 21 minutes, whereas Qwen3-32B completed locally in approximately 14 minutes with no API codes beyond query embedding. Manual UMLS browser collection typically required 1-2 hours of clinician time per concept.

***Table 4. Cost and runtime for CUICurate pipeline setup and execution***. *Embedding generation and FAISS index creation were one-off preprocessing steps performed on the complete UMLS graph. API costs were estimated using the published OpenAI API pricing and the collected input and output token number at each stage (price at the time of execution: text-embedding-3-large input $0.13/million tokens; GPT-5 input $1.25/million tokens, and output $10.00/million tokens). Embedding generation and indexing were performed on a Google Colab L4 GPU instance, and*

*graph retrieval, and GPT-5 CUI filtering and classification were performed using a Google Colab HighRAM CPU instance. Qwen3-32B filtering and classification was performed using a Google Colab A100 GPU instance. Google Colab instances were accessed using a Colab Pro account (additional details provided in the Supplementary materials).*

| Pipeline stage | Method | Approximate Cost (USD) | Approximate Runtime (minutes) |
|---|---|---|---|
| **One-off infrastructure setup** | | | |
| Embedding generation | SapBERT | $0 (run locally) | 16.7 min |
| | e5-large-v2 | $0 (run locally) | 109.5 min |
| | Text-embedding-3-large | $8.20 | 141.2 min |
| FAISS index creation | SapBERT | $0 (run locally) | 2.7 min |
| | e5-large-v2 | $0 (run locally) | 4.9 min |
| | Text-embedding-3-large | $0 (run locally) | 7.6 min |
| **Per-query pipeline execution** | | | |
| Graph retrieval | Text-embedding-3-large query embedding + FAISS retrieval | <$0.01 | 0.14 min |
| LLM filtering | GPT-5 | $0.65 | 16.29 min |
| | Qwen3-32B | $0 (run locally) | 7.96 min |
| LLM classification | GPT-5 | $0.23 | 5.02 min |
| | Qwen3-32B | $0 (run locally) | 5.96 min |

# DISCUSSION

This study demonstrates that combining graph-based retrieval over the UMLS with LLM-guided filtering provides a scalable and reproducible framework to clinical concept set curation. Across five lexically heterogenous target concepts, CUICurate generated significantly larger and more comprehensive concept sets than manual UMLS browser searches, which captured only a small proportion of clinically relevant concepts. A single retrieval configuration using text-embedding-3-large achieved high recall for definitive concepts with manageable candidate sets. GPT-5 consistently outperformed manual curation and preserved nearly all definitive phenotype-defining concepts, while open-source Qwen3-32B achieved comparable, though slightly lower, performance. Many missed concepts were not observed in clinical notes, and computational costs were modest relative to the clinician time required for manual concept set development.

**Graph Retrieval**

Retrieval performance varied substantially across target concepts and was driven primarily by the semantic and ontological characteristics of each rather than by instability of the retrieval method. *CHF* and *LVSD* were retrieved with high recall, reflecting their relatively coherent representation within the UMLS and the presence of well-established terminology. In contrast, FO and PM were more challenging because they are expressed through diverse findings, functional descriptions, and associated conditions rather than a single clearly bounded diagnosis, consistent with previous studies on these phenotypes. [23, 24]

Among the embedding models evaluated, text-embedding-3-large provided the strongest retrieval performance, suggesting that large general-purpose embeddings may outperform domain-specific biomedical models for ontology retrieval tasks. Semantic retrieval alone and unrestricted recursive child expansion rapidly became computationally impractical. CUICurate's combination of semantic retrieval, one-hop expansion and restricted neighbour limits achieved high recall for definitive concepts while maintaining manageable candidate sets for downstream filtering and classification.

For several concepts, particularly FO, PM, and LVSD, the adjudicated gold-standard sets were intentionally broad and included many context-dependent concepts. This reflected the absence of a universally accepted boundary for these phenotypes, supported by lower IAA for broader phenotypes such as FO and PM compared with more clearly defined concepts like IS. The gold-standard sets were therefore designed to test CUICurate's ability to retrieve a wide range of clinically plausible indicators that can be manually reviewed and tailored to specific downstream applications.

**LLM Filtering, Clinical Observability, and Practical Utility**

LLM-based filtering substantially improved concept set completeness compared with manual UMLS browsing while reducing the candidate sets to a more manageable number of relevant CUIs. GPT-5 achieved higher recall than manual concept sets for all five target concepts and preserved at least 95% of definitive gold-standard CUIs, indicating that the most clinically specific concepts were retained with high reliability. Qwen3-32B consistently outperformed manual curation and approached GPT-5 performance, demonstrating that effective concept set generation is feasible using open-source models.

Many of the concepts missed by CUICurate were unlikely to have a substantial impact on downstream clinical text mining. In 10,000 sampled MIMIC-III notes, a large proportion of missed gold-standard CUIs, particularly context-dependent concepts, were not observed in the clinical text, whereas commonly documented definitive concepts were rarely missed. These findings suggest that conventional recall metrics may overstate the practical importance of some omissions.

Qualitative review showed that CUICurate captured numerous clinically meaningful and text-observable concepts absent from manual concept sets. Remaining omissions were largely attributable to predictable and interpretable factors, including prompt-specific exclusions, generic descriptors lacking explicit phenotype specificity, incomplete query coverage, and large families of closely related ontology variants. These errors appear amenable to refinement through targeted prompt engineering and iterative review.

The computational requirements of CUICurate were modest relative to the clinician effort typically required for manual concept set development. With a one-time UMLS embedding cost of less than US$10 and GPT-5 processing cost of less than US$1 per concept, CUICurate offers a cost-efficient approach to automating concept set curation. These low costs enable users to iteratively refine prompts and rerun the pipeline to generate concept sets tailored to specific study objectives.

**Limitations**

Several limitations should be considered. First, the evaluation was restricted to five target concepts. Although, these were intentionally selected to represent lexically heterogeneous phenotypes and concepts with extensive subtypes, performance for other concepts such as procedures, medications, laboratory findings, and rare diseases remains to be established.

Second, despite the use of multiple embedding models, graph expansion strategies, and pooled relevance assessment, some relevant CUIs may remain unretrieved. Pooled relevance assessment provides a practical and rigorous approach to benchmark construction in a search space exceeding 900,000 UMLS concepts but cannot guarantee absolute completeness without exhaustive review.

Third, for concepts such as FO, PM and LVSD, the adjudicated gold-standard sets were intentionally broad and included many context-dependent concepts. This reflects the absence of universally accepted phenotype boundaries and the fact that the concept sets are often tailored to specific analytic objectives. Fourth, the distinction between definitive and context-dependent concepts is partly subjective despite formal adjudication. Fifth, clinical observability was assessed in a single corpus (MIMIC-III) using QuickUMLS and may not generalise to other healthcare settings or named entity recognition tools.

### Future Work

Future work could extend the evaluation to a broader range of concept types, particularly less standardised domains such as nursing and rehabilitation terminologies. CUICurate is best viewed as an iterative concept design tool rather than a fully automated one-step automated solution. Because graph retrieval is rapid and filtering costs are modest, users can inspect retrieved concepts, refine target descriptions and prompts, and rerun the pipeline to generate study-specific concept sets tailored to different analytical objectives and balances between sensitivity and specificity. This process may be further improved by incorporating richer concept descriptions, including clinical diagnostic criteria and relevant clinical literature. Additional engineering work may reduce runtime through parallel processing and optimised prompting. Finally, downstream evaluations in concept normalisation, cohort definition, and automated registry creation, will be important to establish their clinical and research value of CUICurate-generated concept sets, which will be explored in future studies.

## CONCLUSION

CUICurate, is a scalable and reproducible framework for automated UMLS concept set curation that combines graph-based retrieval with LLM-guided filtering and classification. Across five clinical concepts, it consistently outperformed manual UMLS browser searches, generating larger and more complete concept sets while maintaining acceptable precision.

The findings show that concept set curation can benefit from a GraphRAG framework in which graph-based retrieval constrains the search space, and LLMs prioritise clinically relevant concepts for expert review. By providing a practical and cost-efficient approach to generating clinician-reviewable, study-specific concept sets, CUICurate supports more consistent and maintainable use of UMLS-based concept normalisation in clinical NLP and phenotyping applications.

**ACKNOWLEDGMENTS**

We thank the clinicians who contributed their time to the manual concept collection and adjudication. Generative AI tools (ChatGPT, OpenAI) were used to assist with language editing, restructuring of text for clarity, and debugging of Python analysis scripts. AI tools were not used to generate data, perform analyses, or interpret results. All outputs were reviewed, validated, and revised by the authors, who take full responsibility for the content.

**AUTHOR CONTRIBUTIONS**

VB designed the study, conducted all experiments, performed the analyses and drafted the manuscript. MM, JN and VB conducted the clinician-led concept collection and adjudication. BG and SO supervised the project and provided methodological and clinical guidance. All authors reviewed and approved the final manuscript.

**FUNDING**

VB is supported by an Australian Government Research Training Program (RTP) Scholarship and an Industry PhD Scholarship, including stipend support from Eastern Heart Clinic.

**CONFLICT OF INTEREST**

The authors declare no conflicts of interest.

## ETHICS STATEMENT

This study did not involve human participants. Access to the MIMIC-III database was obtained through PhysioNet after completion of the required human research training and acceptance of the associated Data Use Agreement.

## DATA AVAILABILITY

The CUICurate framework is available at https://github.com/vickyblake/CUICurate. The gold-standard concept sets created by this project are provided in the Supplementary materials, Unified Medical Language System data tables are distributed by the U.S. National Library of Medicine under licence and are not publicly sharable. Researchers can apply for a free UMLS Metathesaurus licence directly through the National Library of Medicine at https://www.nlm.nih.gov/research/umls/index.html**.** MIMIC-III data can be accessed through PhysioNet here https://physionet.org/content/mimiciii/1.4/

# REFERENCES

1. Xu H, Demner Fushman D, Hong N, Raja K. Medical Concept Normalization. In: Xu H, Demner Fushman D, editors. *Natural Language Processing in Biomedicine: A Practical Guide*. Cham: Springer International Publishing; 2024. p. 137-64.

2. Jing X. The Unified Medical Language System at 30 Years and How It Is Used and Published: Systematic Review and Content Analysis. *JMIR Med Inform* 2021; **9**: e20675.

3. Gold S, Batch A, McClure R, et al. Clinical Concept Value Sets and Interoperability in Health Data Analytics. *AMIA Annu Symp Proc* 2018; **2018**: 480-9.

4. Soldaini L, Goharian N. *Quickumls: a fast, unsupervised approach for medical concept extraction*. MedIR workshop, sigir; 2016; 2016. p. 1-4.

5. Kraljevic Z, Bean D, Mascio A, et al. MedCAT -- Medical Concept Annotation Tool. *arXiv pre-print server* 2019.

6. Rodriguez VA, Tony S, Thangaraj P, et al. Phenotype Concept Set Construction from Concept Pair Likelihoods. *AMIA Annu Symp Proc* 2020; **2020**: 1080-9.

7. *UMLS Metathesaurus Browser*. [cited 28/09/2025]; Available from: https://uts.nlm.nih.gov/uts/umls

8. *Athena Search Terms*. [cited 28/09/2025]; Available from: https://athena.ohdsi.org/search-terms/start

9. Kipp M. From GPT-3.5 to GPT-4.o: A Leap in AI's Medical Exam Performance. *Information* 2024; **15**: 543.

10. Lewis P, Perez E, Piktus A, et al. Retrieval-augmented generation for knowledge-intensive NLP tasks. *Proceedings of the 34th International Conference on Neural Information Processing Systems*. Vancouver, BC, Canada: Curran Associates Inc.; 2020. p. Article 793.

11. Han H, Wang Y, Shomer H, et al. Retrieval-augmented generation with graphs (graphrag). *arXiv preprint arXiv:250100309* 2024.

12. Lukyanchikov N, Kawamoto K. Evaluation of Discrepancies Among National Library of Medicine (NLM) Value Set Authority Center (VSAC) ICD-10-CM Value Sets: Case Study for Diagnoses of Common Chronic Conditions, Implications, and Potential Solutions. *AMIA Annu Symp Proc* 2023; **2023**: 1087-95.

13. Dobbins NJ. Generalizable and scalable multistage biomedical concept normalization leveraging large language models. *Research Synthesis Methods* 2025; **16**: 479-90.

14. Banf M, Kuhn J. A Tripartite Perspective on GraphRAG. *arXiv preprint arXiv:250419667* 2025.

15. Goel S, Lee RJ, Ramchandran K. SAGE: A Realistic Benchmark for Semantic Understanding. *arXiv preprint arXiv:250921310* 2025.

16. Liu F, Shareghi E, Meng Z, Basaldella M, Collier N. Self-alignment pretraining for biomedical entity representations. *arXiv preprint arXiv:201011784* 2020.

17. Excoffier J-B, Roehr T, Figueroa A, Papaioannou J-M, Bressem K, Ortala M. Generalist embedding models are better at short-context clinical semantic search than specialized embedding models. *arXiv preprint arXiv:240101943* 2024.

18. Wang L, Yang N, Huang X, et al. Text embeddings by weakly-supervised contrastive pre-training. *arXiv preprint arXiv:221203533* 2022.

19. Douze M, Guzhva A, Deng C, et al. The faiss library. *arXiv preprint arXiv:240108281* 2024.

20. Yang A, Li A, Yang B, et al. Qwen3 technical report. *arXiv preprint arXiv:250509388* 2025.

21. Sanderson M. Test collection based evaluation of information retrieval systems. *Foundations and Trends® in Information Retrieval* 2010; **4**: 247-375.

22. Voorhees EM, Soboroff I, Lin J. Can old TREC collections reliably evaluate modern neural retrieval models? *arXiv preprint arXiv:220111086* 2022.

23. Thieu T, Maldonado JC, Ho P-S, et al. A comprehensive study of mobility functioning information in clinical notes: entity hierarchy, corpus annotation, and sequence labeling. *International journal of medical informatics* 2021; **147**: 104351.

24. Messmer AS, Moser M, Zuercher P, Schefold JC, Müller M, Pfortmueller CA. Fluid Overload Phenotypes in Critical Illness—A Machine Learning Approach. *Journal of Clinical Medicine*; 2022. p. 336.

# Supplementary Material

# Computing environment

Google Colab with a Google Colab Pro account was used for all framework development and testing . The graphRAG components (graph, embeddings and indexes) were built using an L4 GPU instance type, the GPT-5 experiments were run using a high RAM CPU instance, accessing the model through the OpenAI API. The Qwen3-32B experiments were run using an A100 high RAM GPU instance.

# Manual and Gold-Standard Concept Set Agreement Statistics

The manually collected concept sets ***(M)*** showed low overlap between the annotators (Jaccard = 0.08-0.37; overlap coefficient = 0.19-0.65) due to the very large CUI search space and differing search paths within the UMLS Metathesaurus browser. Agreement between the two annotators for the gold-standard ($G_a$) manually adjudicated concept sets was generally high for inclusion decisions (percent agreement = 83.9-92.6%; $\kappa$ = 0.64-0.69) across concepts. Category agreement CUIs was more variable (percent agreement 67.5-94.1%; $\kappa$ = 0.32-0.84).

Supplementary Table 1. Overlap between manually collected UMLS browser concept sets (M).

| | TOTAL CUIS | INTERSECTION SIZE | JACCARD | OVERLAP COEFFICIENT |
|---|---|---|---|---|
| Fluid Overload | 41 | 15 | 0.37 | 0.65 |
| Ischaemic Stroke | 166 | 24 | 0.15 | 0.27 |
| LV Systolic Dysfunction | 68 | 12 | 0.18 | 0.31 |
| Poor Mobility | 148 | 12 | 0.08 | 0.19 |

Supplementary Table 2. Inter-annotator agreement for gold-standard (Ga) concept sets

| Metric | Chronic heart failure | Fluid overload | Ischaemic stroke | LVSD | Poor Mobility |
|---|---|---|---|---|---|
| Total CUIs in pooled superset (n) | 1835 | 2071 | 1662 | 2111 | 2140 |
| Annotator 1 included (n, %) | 161 (8.8%) | 292 (14.1%) | 674 (40.6%) | 264 (12.5%) | 779 (36.4%) |
| Annotator 2 included (n, %) | 410 (22.3%) | 205 (9.9%) | 619 (37.2%) | 291 (13.8%) | 512 (23.9%) |
| Include Agreement (%) | 82.3% | 90.2% | 93.9% | 94.5% | 73.7% |
| Include Cohen's Kappa | 0.349 | 0.542 | 0.872 | 0.757 | 0.387 |
| Inclusion conflicts (n) | 325 | 202 | 101 | 117 | 563 |
| Category Agreement (%) | 67.5% | 82.4% | 85.7% | 75.3% | 73.6% |
| Category Cohen's Kappa | 0.306 | 0.519 | 0.664 | 0.467 | 0.431 |
| Category conflicts (n) | 40 | 26 | 85 | 54 | 96 |
| Final included CUIs (n, %) | 163 (8.9%) | 321 (15.5%) | 675 (40.6%) | 287 (13.6%) | 655 (30.6%) |
| Final definitive CUIs (n, %) | 114 (69.9%) | 23 (7.2%) | 403 (59.7%) | 51 (17.8%) | 355 (54.2%) |
| Final context-dependent CUIs (n, %) | 49 (30.1%) | 298 (92.8%) | 273 (40.4%) | 236 (82.2%) | 300 (45.8%) |

# Graph Retrieval Settings and Additional Experiments

Supplementary Table 3. Definitions of graph retrieval settings used to limit the graph retrieval of candidate CUIs to control prompt noise and length

| **Retrieval Setting** | **Description** |
|---|---|
| Target CUI | The CUI with the top semantic similarity to the target concept description |
| K | Top-*k* most similar CUIs as initial seeds |
| Children | Descendent nodes of the seed CUIs |
| Hops | Number of edge traversals from the seed CUIs (e.g. synonyms and broader terms that are not descendants of K) |
| Max Neighbours | Maximum threshold number of nearest neighbour CUIs retrieved from the neighbourhood of the target CUI (limited prompt length and reduced noise) |
| Semantic Type | UMLS semantic types eligible for retrieval |

Supplementary Table 4. Semantic retrieval only strategy at various bands of target recall for the five target concepts using the adjudicated gold-standard concept sets. The number of CUIs retrieved in the candidate concept sets quickly escalates for higher target recall values, particularly for fluid overload (FO) and poor mobility (PM).

| | **CHF** | | **FO** | | **LVSD** | | **IS** | | **PM** | |
|---|---|---|---|---|---|---|---|---|---|---|
| **Target recall** | **CUIs retrieved** | **Gold missed** | **CUIs retrieved** | **Gold missed** | **CUIs retrieved** | **Gold missed** | **CUIs retrieved** | **Gold missed** | **CUIs retrieved** | **Gold missed** |
| 0.60 | 138 | 64 | 591 | 134 | 328 | 99 | 422 | 235 | 499 | 176 |
| 0.70 | 179 | 48 | 812 | 101 | 411 | 74 | 518 | 176 | 695 | 132 |
| 0.80 | 272 | 32 | 1179 | 67 | 573 | 49 | 646 | 117 | 1084 | 88 |
| 0.85 | 329 | 24 | 2029 | 50 | 718 | 37 | 708 | 88 | 1805 | 66 |
| 0.90 | 382 | 16 | 4810 | 33 | 921 | 24 | 848 | 58 | 3060 | 44 |
| 0.95 | 505 | 8 | 12453 | 16 | 1398 | 12 | 1110 | 29 | 7220 | 22 |
| 1.00 | 763 | 0 | 60426 | 0 | 4022 | 0 | 5006 | 0 | 32967 | 0 |

# Target Concept Query Optimisation

To optimise the recall of the retrieval query for each target concept, GPT-5-mini was used to craft descriptions of the target concept using the below prompt. The 3 LLM-generated descriptions were then reviewed by a clinician and combined to form one description that covered the key synonyms, subtypes and variant clinical phrases that would indicate the target concept.

```
You are a clinical NLP expert optimising retrieval recall for UMLS GraphRAG.
 Target concept: {concept}

 Write 3 alternative, recall-oriented descriptions (<=120 words each) to use as dense retrieval queries.
 Each description MUST:
      - Include synonyms and lexical variants (incl. US/UK spellings)
      - Include common abbreviations if applicable
      - Mention key indicators (specific signs, symptoms, measurements, labs, procedures, and subtypes)
      - Avoid broad, non-specific symptoms
      - Be one paragraph per variant
```

# Target Concept Description Queries

Supplementary Table 5. Target concept description text used as queries for CUI retrieval, semantic types and additional LLM filtering instrictions and classification few-shots. Concept descriptions embedded and used to calculate cosine similarity to UMLS graph nodes and guide LLM prompt filtering and classification of CUIs

| **Target Concept** | **Concept Description Query Embedded and** | **Semantic Types** | **LLM filtering instructions** | **LLM classification few-shots** |
|---|---|---|---|---|
| Chronic heart failure | Synonyms, equivalent terms and abbreviations:<br>Chronic heart failure.<br>Chronic cardiac failure,<br>congestive heart failure,<br>chronic congestive cardiac failure,<br>chronic cardiac insufficiency,<br>pump failure<br><br>Subtypes and variants:<br>Heart failure with reduced ejection fraction (HFrEF)<br>Heart failure with preserved ejection fraction (HFpEF)<br>Heart failure with mid-range ejection fraction (HFmrEF)<br>systolic heart failure<br>diastolic heart failure,<br>right-sided heart failure,<br>left-sided heart failure,<br>bilventricular heart failure. | Disease or Syndrome,<br>Pathologic Function,<br>Diagnostic Procedure,<br>Health Care Activity,<br>Finding,<br>Laboratory or Test Result | Include:<br>- "heart failure" with unspecified acuity, “acute-on-chronic” presentations and compensated or decompensated states.<br>- concepts for heart failure hospitalisations or care events.<br><br>Exclude:<br>- concepts related to acute cardiac failure (e.g. cardiogenic shock) that are are likely acute issues that may resolve, for example sepsis, post-procedure.<br>- concepts related to fetal or neonate cardiac failure. | definitive: chronic congestive cardiac failure, heart failure with reduced ejection fraction<br><br>context dependent: heart failure, congestive cardiac failure |
| | Objective indicators:<br>Ventricular dysfunction.<br>Reduced LVEF on echocardiography<br>Cardiomyopathy<br>Exertional dyspnoea / dyspnea<br>Orthopnoea / orthopnea<br>Paroxysmal nocturnal dyspnoea / dyspnea<br>Elevated BNP or NT-proBNP<br>Cardiomegaly or pulmonary congestion on chest X-ray<br>New York Heart Association score (NYHA) | | | |
| Fluid overload | Synonyms, equivalent terms and abbreviations:<br>Systematic fluid overload<br>Systemic volume overload,<br>hypervolemia,<br>fluid retention,<br>fluid excess,<br>circulatory overload,<br>volume expansion.<br>volume overload<br><br>Subtypes and variants:<br>Pulmonary edema/ pulmonary oedema<br>Dependent (peripheral) edema<br>Anasarca (generalised edema)<br>Transfusion-associated circulatory overload (TACO)<br>Iatrogenic volume overload | Disease or Syndrome,<br>Pathologic Function,<br>Diagnostic Procedure,<br>Health Care Activity,<br>Finding,<br>Laboratory or Test Result,<br>Sign or Symptom | Include:<br>- concepts for hospitalisations or care events related to fluid overload.<br><br>Exclude:<br>- concepts related to fetal fluid overload. | definitive: pulmonary oedema due to fluid overload, hypervolaemia<br><br>context dependent: oedema of the lower limbs, fluid imbalance |

|  | Renal failure-associated volume overload<br>Cardiogenic volume overload.<br><br>Objective indicators:<br>Positive fluid balance<br>acute weight gain,<br>elevated central venous pressure (CVP),<br>Jugular venous distention (elevated JVP)<br>Generalised oedema<br>Generalised edema<br>dilated non-collapsible Inferior Vena Cava,<br>Vascular congestion<br>elevated B-type natriuretic peptide (BNP)<br>elevated N-terminal pro-B-type natriuretic peptide (NT-proBNP),<br>Kerley B lines,<br>Interstitial edema<br>Alveolar edema<br>Pleural effusion,<br>requirement for dialysis<br>requirement for ultrafiltration,<br>response to loop diuretics<br>response to ultrafiltration,<br>Bilateral peripheral edema<br>Bilateral pitting edema,<br>Ascites |  |  |  |
|---|---|---|---|---|
| Left ventricular systolic dysfunction | Synonyms, equivalent terms and abbreviations:<br>LV systolic dysfunction,<br>Left ventricular dysfunction<br>Left ventricular systolic failure,<br>systolic heart failure,<br>heart failure with reduced ejection fraction, HFrEF,<br>impaired left ventricular contractility<br>acute systolic dysfunction<br>chronic systolic dysfunction<br><br>Subtypes and variants:<br>Dilated cardiomyopathy<br>post-MI/ischaemic (ischemic) cardiomyopathy<br>Post-MI cardiomyopathy<br>Ischaemic cardiomyopathy<br>non-ischaemic cardiomyopathy,<br>myocarditis-related cardiomyopathy,<br>anthracycline induced cardiomyopathy<br>toxin-induced cardiomyopathy<br>post-infarct remodelling,<br>cardiomegaly.<br><br>Objective indicators:<br>low stroke volume<br>cardiac output,<br>depressed LVEF/EF (eg EF <40% or <35%),<br>reduced ejection fraction | Disease or Syndrome,<br>Pathologic Function,<br>Diagnostic Procedure,<br>Health Care Activity,<br>Finding,<br>Laboratory or Test Result, | Include:<br>- concepts for hospitalisations or care events related to Left Ventricular Systolic Dysfunction.<br><br>Exclude:<br>- concepts related to foetal ventricular function.<br>- concepts related to clinical manifestations of systolic dysfunction like pulmonary oedema without the concept specifying that it is related to LV dysfunction. | definitive: left-sided systolic failure, heart failure with reduced ejection fraction<br><br>context dependent: dilated cardiomyopathy, ventricle wall-motion abnormality |
| | reduced fractional shortening,<br>global hypokinesis<br>regional wall-motion abnormality<br>abnormal transthoracic/TEE echocardiography,<br>pulmonary oedema/edema,<br>raised pulmonary capillary wedge pressure,<br>low cardiac index (<2.2 L/min/m2),<br>elevated BNP<br>elevated NT-proBNP,<br>S3 gallop | | | |
| Ischaemic Stroke | Synonyms, equivalent terms and abbreviations:<br>stroke,<br>ischemic stroke<br>cerebrovascular accident (CVA),<br>cerebral infarction,<br>acute brain infarct,<br>brain ischemia.<br>Brain ischaemia<br><br>Subtypes and variants:<br>Thromboembolic stroke<br>thrombotic infarct,<br>embolic infarct,<br>cardioembolic infarct<br>lacunar infarct,<br>small-vessel infarct<br>watershed infarct<br>large vessel occlusion (LVO) infarct. | Disease or Syndrome,<br>Pathologic Function,<br>Diagnostic Procedure,<br>Health Care Activity,<br>Finding,<br>Laboratory or Test Result, | Include:<br>- concepts for hospitalisations or care events related to ischaemic stroke.<br><br>Exclude:<br>- concepts related to foetal. | definitive: cerebral infarction, MCA/ICA territory infarct<br><br>context dependent: acute cerebrovascular accident (CVA), stroke |
| | Large artery atherosclerotic stroke (TOAST)<br>Middle Cerebral Artery infarct<br>Internal Carotid Artery territory infarct,<br>Anterior Cerebral Artery infarct,<br>Posterior Cerebral Artery infarct,<br>basal ganglia infarct,<br>brainstem infarct,<br>cerebellum infarct.<br>Cryptogenic stroke<br><br>Objective indicators:<br>sudden focal deficits,<br>unilateral weakness<br>unilateral hemiparesis,<br>sensory lacunar syndromes,<br>cortical signs,<br>facial droop,<br>sudden aphasia (expressive/global),<br>expressive aphasia<br>global aphasia<br>dysarthria,<br>neglect,<br>homonymous hemianopia,<br>ataxia<br>brainstem signs<br>diplopia,<br>dysphagia<br>National Institute of Health Stroke score (NIHSS).<br>CTB hypodensity, | | | |

| | CTB perfusion mismatch,<br>CTB occlusion<br>MRA brain occlusion<br>Carotid stenosis<br>DWI MRI diffusion restriction,<br>Intravenous thrombolysis.<br>mechanical thrombectomy (MT),<br>endovascular thrombectomy (EVT). | | | |
|---|---|---|---|---|
| Poor Mobility | Synonyms, equivalent terms and abbreviations:<br>poor mobilization/mobilisation;<br>reduced mobility;<br>decreased mobility<br>difficult ambulation<br>impaired mobility<br>limited mobility<br>mobility limitation<br>non-ambulatory;<br>difficulty ambulating<br>immobilised/immobilized<br><br>subtypes and variants:<br>Gait disturbance / Gait disorder / Impaired gait<br>Balance impairment / postural instability<br>Reduced independent ambulation<br>Ataxic gait.<br>Wheelchair-dependent mobility<br>Difficulty walking.<br>Difficulty ambulating. | Disease or Syndrome,<br>Pathologic Function,<br>Diagnostic Procedure,<br>Health Care Activity,<br>Finding,<br>Laboratory or Test Result, | Include:<br>- concepts for hospitalisations or care events related to Poor Mobility.<br>- concepts related to use of mobility aids.<br><br>Exclude:<br>- concepts related to the measurement of mobility without any assertion that the mobility is limited or poor.<br>- concepts that are Conditions or impairments associated with poor mobility where mobility impairment is not explicit or inevitable | definitive: difficulty walking, use of walking stick, use of wheelchair<br><br>context dependent: recurrent falls, ataxia, difficulty kneeling, decreased range of movement in lower body region |
| | Does not walk.<br>Does not mobilise.<br>Transfer impairment.<br>Sit-to-stand impairment.<br>Wheelchair-bound.<br>Bed-bound.<br><br>Objective indicators:<br>Transfer dependence.<br>Assistance-dependent mobilisation.<br>Uses walking stick.<br>Uses cane.<br>Uses walker.<br>Uses rollator.<br>Uses wheelchair.<br>Slow gait speed.<br>Abnormal tandem or unsteady gait<br>Timed Up and Go test > 12 seconds.<br>Reduced six-minute walk test distance (6MWT).<br>Reduced 10-metre walk performance.<br>Inability to perform five times sit-to-stand (STS).<br>Recurrent falls. | | (e.g. early stages of Parkinson's disease) | |

# LLM Prompting

## LLM Filtering Prompt Template

```
You are a biomedical assistant working with a UMLS-based knowledge graph.

Task: Filter the CUIs provided below that should reasonably be considered to indicate the target concept.

Include if:
    1) It directly denotes the target concept or is a clinically equivalent synonym.
    2) It is a subtype/child that is commonly subsumed by the target concept without additional qualifiers.
    3) It is a closely bound variant naming (spelling variants, common aliases).
    4) It is a procedure, therapy, medication, lab/test, measurement, risk factor, aetiology, complication,
       manifestation, generic finding, or care event (e.g., specialty clinic attendance, disease-management
programs, surgery admissions) that is widely recognised as a proxy or near-unique indicator of the target
concept.

Exclude if:
    1) It is a generic parent/container concept that is not commonly used as the disease name.
    2) It expresses only suspicion, family history, or negation of the target concept.
    3) It is clearly unrelated or ambiguous.

Prefer inclusion to preserve recall for direct assertion concepts. For conditional categories in 4),
include only when specificity is high; if uncertain about specificity, exclude.

Target concept: {concept_name} (CUI: {concept_cui})

Target concept description/aliases: {target_description}

Special include/exclude instructions: {special_instructions}

OUTPUT FORMAT (strict):
     - Return ONLY valid JSON with exactly one key: "selected_cuis".
     - "selected_cuis" MUST be an array of CUIs (strings matching ^C\\d{7}$).
```

```
     - You may ONLY output CUIs from the provided candidate list below.
     - No prose, no comments, no extra keys.
     - De-duplicate and sort CUIs ascending for stability.

Candidates:
        {cui}: {cui_name}
        {cui}: {cui_name}
        {cui}: {cui_name}
```

## LLM Classification Prompt Template

```
You are a biomedical assistant working with UMLS CUIs.

Task: Classify every provided CUI listed below into exactly one of the categories as defined below.

Rules:
    1) "definitive": direct, unambiguous assertion of the target concept (or strict synonyms).
    2) "context_dependent": modifiers, tests, measurements, risk factors, causes, complications,
manifestations: include parents only if the term commonly used to refer to the target concept; subtypes that
need qualifiers.
    3) Tie-breakers: when uncertain, prefer "context_dependent"; choose specific assertion over generic
parent.
    4) No omissions, no duplicates.

Target concept: {target_description}

Few-shot examples: {fewshots}

Return ONLY valid JSON with exactly these keys: "definitive" and "context_dependent".
Values MUST be arrays of CUIs (strings like "C1234567") drawn ONLY from the provided list.
Do NOT include names, comments, or extra keys.
   Example:
       {
              "definitive": ["C0000001", "C0000002"],
              "context_dependent": ["C0000003"]
```

```
        }

CUIs:
        {cui}: {cui_name}
        {cui}: {cui_name}
        {cui}: {cui_name}
```

# Graph Retrieval Performance

## Evaluation Framework

To evaluate the automated system, each stage of the pipeline was assessed separately. All evaluations were conducted on the CUIs accessible to each method (i.e., the method's retrieval or inclusion space) to ensure fair comparisons.

1. **Graph Retrieval Performance**

   Graph retrieval (***R***) performance was evaluated against the clinician benchmark set ***M.*** Recall was the primary optimisation target and was calculated as:

$$Recall_{retrieval} = \frac{|M \cap R|}{|M|}$$

2. **Filtering Performance**

   The LLM filtering step and the manual benchmark were evaluated against the adjudicated gold-standard $G_a$, which comprises the pooled CUIs retrieved from three embedding models. These CUIs were independently reviewed and annotated by clinicians with consensus adjudication.

   **Evaluation of LLMs**

LLM predictions ($\boldsymbol{P^{llm}}$) were compared with the gold-standard $\boldsymbol{G_a}$. Evaluation was restricted to CUIs retrievable under GraphRAG (i.e., $\boldsymbol{G_a} \cap \boldsymbol{R}$), matching the LLMs' search space.

$$Recall_{llm} = \frac{|\ G_a \cap P^{llm}\ |}{|\ G_a \cap R\ |}$$

$$Precision_{llm} = \frac{|\ G_a \cap P^{llm}\ |}{|\ P^{llm}\ |}$$

**Evaluation of the Manual Benchmark**

Because $\boldsymbol{G_a}$ is constrained by GraphRAG retrieval, it may omit valid CUIs collected by clinicians simply because retrieval did not surface them. Treating such CUIs as false positives would therefore misrepresent manual performance.

To ensure a fair comparison, all CUIs proposed by clinicians in the manual concept sets ($\boldsymbol{P^M}$) were treated as condition-positive for inclusion evaluation. Manual performance therefore focuses on recall of the adjudicated gold:

$$Recall_{M} = \frac{|\ G_a \cap P^{M}\ |}{|\ G_a\ |}$$

## 3. Definitive CUI Performance

Definitive CUI performance was evaluated only for CUIs included during the preceding inclusion stage for each method. Each CUI, was evaluated against with the gold-standard classification in $G_a$, and precision, recall, and F1-scores were computed.

Supplementary Table 6. Semantic retrieval only results. Number of CUIs retrieved and number of gold-standard CUIs missed at various recall values

| Target recall | CHF | | FO | | LVSD | | IS | | PM | |
|---|---|---|---|---|---|---|---|---|---|---|
| | CUIs retrieved | Gold missed | CUIs retrieved | Gold missed | CUIs retrieved | Gold missed | CUIs retrieved | Gold missed | CUIs retrieved | Gold missed |
| 0.6 | 138 | 64 | 591 | 134 | 328 | 99 | 422 | 235 | 499 | 176 |
| 0.7 | 179 | 48 | 812 | 101 | 411 | 74 | 518 | 176 | 695 | 132 |
| 0.8 | 272 | 32 | 1179 | 67 | 573 | 49 | 646 | 117 | 1084 | 88 |
| 0.85 | 329 | 24 | 2029 | 50 | 718 | 37 | 708 | 88 | 1805 | 66 |
| 0.9 | 382 | 16 | 4810 | 33 | 921 | 24 | 848 | 58 | 3060 | 44 |
| 0.95 | 505 | 8 | 12453 | 16 | 1398 | 12 | 1110 | 29 | 7220 | 22 |
| 1.0 | 763 | 0 | 60426 | 0 | 4022 | 0 | 5006 | 0 | 32967 | 0 |

Supplementary Table 7. Retrieval results at the selected graph retrieval settings for five target concepts.

| concept | model | k | max_neighbors | hops | retrieved_n | recall | precision | f1 | recall_def |
|---|---|---|---|---|---|---|---|---|---|
| Chronic Heart Failure | text-embedding-3-large | 500 | 1000 | 1 | 758 | 0.962 | 0.204 | 0.337 | 0.960 |
| Fluid Overload | text-embedding-3-large | 500 | 1000 | 1 | 768 | 0.721 | 0.316 | 0.439 | 0.961 |
| Ischaemic Stroke | text-embedding-3-large | 500 | 1000 | 1 | 764 | 0.857 | 0.660 | 0.746 | 0.900 |
| LV Systolic Dysfunction | text-embedding-3-large | 500 | 1000 | 1 | 743 | 0.895 | 0.298 | 0.448 | 0.925 |
| Poor Mobility | text-embedding-3-large | 500 | 1000 | 1 | 782 | 0.666 | 0.409 | 0.507 | 0.765 |

## LLM Filtering Performance

Supplementary Table 8. LLM filtering performance. Number of CUIs included in each concept set, with mean recall, precision and F1-score across five runs per concept for GPT-5 and Qwen3-32B, and compared with the manual UMLS browser collected concept sets. Each concept set was evaluated against the adjudicated gold-standard CUIs.

| **Evaluation Type** | **Concept** | **Method** | **Gold CUIs, n** | **Predicted CUIs, n** | **Recall** | **Precision** | **F1** |
|---|---|---|---|---|---|---|---|
| All CUIs | CHF | Manual | 163 | 98 | 0.60 | 1.00 | 0.75 |
| | | GPT-5 | 157 | 193 (2) | 0.82 (0.02) | 0.66 (0.01) | 0.73 (0.01) |
| | | Qwen3-32B | 157 | 178 (7) | 0.69 (0.03) | 0.61 (0.00) | 0.65 (0.02) |
| | FO | Manual | 325 | 30 | 0.09 | 1.00 | 0.17 |
| | | GPT-5 | 238 | 86 (4) | 0.33 (0.01) | 0.91 (0.02) | 0.48 (0.02) |
| | | Qwen3-32B | 238 | 48 (8) | 0.17 (0.02) | 0.83 (0.06) | 0.28 (0.03) |
| | IS | Manual | 702 | 130 | 0.19 | 1.00 | 0.31 |
| | | GPT-5 | 532 | 423 (6) | 0.78 (0.01) | 0.98 (0.00) | 0.87 (0.01) |
| | | Qwen3-32B | 532 | 401 (0) | 0.69 (0.00) | 0.92 (0.00) | 0.79 (0.00) |
| | LVSD | Manual | 290 | 55 | 0.19 | 1.00 | 0.32 |
| | | GPT-5 | 243 | 175 (5) | 0.64 (0.01) | 0.89 (0.02) | 0.74 (0.01) |
| | | Qwen3-32B | 243 | 80 (0) | 0.30 (0.00) | 0.91 (0.00) | 0.45 (0.00) |
| | PM | Manual | 677 | 92 | 0.14 | 1.00 | 0.24 |
| | | GPT-5 | 390 | 372 (15) | 0.77 (0.02) | 0.81 (0.02) | 0.79 (0.01) |
| | | Qwen3-32B | 390 | 352 (0) | 0.66 (0.00) | 0.73 (0.00) | 0.70 (0.00) |
| Definitive CUIs | CHF | Manual | 114 | 98 | 0.79 | 0.92 | 0.85 |
| | | GPT-5 | 110 | 193 (2) | 0.96 (0.01) | 0.55 (0.01) | 0.70 (0.01) |
| | | Qwen3-32B | 110 | 178 (7) | 0.76 (0.04) | 0.47 (0.01) | 0.58 (0.02) |
| | FO | Manual | 23 | 30 | 0.61 | 0.47 | 0.53 |
| | | GPT-5 | 23 | 86 (4) | 0.97 (0.04) | 0.26 (0.01) | 0.41 (0.01) |
| | | Qwen3-32B | 23 | 48 (8) | 0.77 (0.02) | 0.38 (0.08) | 0.51 (0.06) |
| | IS | Manual | 403 | 130 | 0.17 | 0.54 | 0.26 |
| | | GPT-5 | 335 | 423 (6) | 0.95 (0.01) | 0.75 (0.01) | 0.84 (0.01) |
| | | Qwen3-32B | 335 | 401 (0) | 0.80 (0.00) | 0.67 (0.00) | 0.73 (0.00) |
| | LVSD | Manual | 51 | 55 | 0.55 | 0.51 | 0.53 |
| | | GPT-5 | 51 | 175 (5) | 0.95 (0.01) | 0.28 (0.01) | 0.43 (0.01) |
| | | Qwen3-32B | 51 | 80 (0) | 0.65 (0.00) | 0.41 (0.00) | 0.50 (0.00) |
| | PM | Manual | 355 (0) | 92 | 0.17 (0.00) | 0.64 (0.00) | 0.26 (0.00) |
| | | GPT-5 | 236 (0) | 372 (15) | 0.95 (0.01) | 0.60 (0.02) | 0.73 (0.02) |
| | | Qwen-32B | 236 (0) | 352 (0) | 0.76 (0.00) | 0.51 (0.00) | 0.61 (0.00) |